\documentclass[runningheads]{llncs}

\usepackage{graphicx}
\usepackage{tikz}
\usepackage{comment}
\usepackage{amsmath,amssymb} % define this before the line numbering.
\usepackage{color}

\usepackage[accsupp]{axessibility}  % Improves PDF readability for those with disabilities.

\usepackage{booktabs}
\usepackage{multirow}
\usepackage[caption=false]{subfig}
\usepackage[para]{footmisc}
\usepackage[pagebackref,breaklinks,colorlinks]{hyperref}

\usepackage[capitalize]{cleveref}
\crefname{section}{Sec.}{Secs.}
\Crefname{section}{Section}{Sections}
\Crefname{table}{Table}{Tables}
\crefname{table}{Tab.}{Tabs.}

\newcommand*\samethanks[1][\value{footnote}]{\footnotemark[#1]}

\begin{document}
\pagestyle{headings}
\mainmatter

\def\ECCVSubNumber{1740}  % Insert your submission number here

\title{SepLUT: Separable Image-adaptive Lookup Tables for Real-time Image Enhancement} % Replace with your title

\begin{comment}
\titlerunning{ECCV-22 submission ID \ECCVSubNumber} 
\authorrunning{ECCV-22 submission ID \ECCVSubNumber} 
\author{Anonymous ECCV submission}
\institute{Paper ID \ECCVSubNumber}
\end{comment}
%******************

% \begin{comment}
\titlerunning{SepLUT}
\author{Canqian Yang\inst{1}\thanks{Equal Contribution}
\quad
Meiguang Jin\inst{2}\samethanks[1]
\quad
Yi Xu\inst{1}\thanks{Corresponding Author\newline \hspace*{1em}Work partially done during an internship of C. Yang at Alibaba Group.}
\quad
Rui Zhang\inst{1}
\\
Ying Chen\inst{2}
\quad
Huaida Liu\inst{2}
}
\authorrunning{C. Yang et al.}
\institute{MoE Key Lab of Artificial Intelligence, AI Institute, Shanghai Jiao Tong University\\
\email{\{charles.young, xuyi, zhang\_rui\}@sjtu.edu.cn}
\and
Alibaba Group\\
\email{\{meiguang.jmg, yingchen, liuhuaida.lhd\}@alibaba-inc.com}
}
% \end{comment}
%******************

\maketitle

%%%%%%%%% ABSTRACT
\begin{abstract}

Image-adaptive lookup tables (LUTs) have achieved great success in real-time image enhancement tasks due to their high efficiency for modeling color transforms. However, they embed the complete transform, including the color component-independent and the component-correlated parts, into only a single type of LUTs, either 1D or 3D, in a coupled manner. This scheme raises a dilemma of improving model expressiveness or efficiency due to two factors. On the one hand, the 1D LUTs provide high computational efficiency but lack the critical capability of color components interaction. On the other, the 3D LUTs present enhanced component-correlated transform capability but suffer from heavy memory footprint, high training difficulty, and limited cell utilization. Inspired by the conventional divide-and-conquer practice in the image signal processor, we present SepLUT (separable image-adaptive lookup table) to tackle the above limitations. Specifically, we separate a single color transform into a cascade of component-independent and component-correlated sub-transforms instantiated as 1D and 3D LUTs, respectively. In this way, the capabilities of two sub-transforms can facilitate each other, where the 3D LUT complements the ability to mix up color components, and the 1D LUT redistributes the input colors to increase the cell utilization of the 3D LUT and thus enable the use of a more lightweight 3D LUT. Experiments demonstrate that the proposed method presents enhanced performance on photo retouching benchmark datasets than the current state-of-the-art and achieves real-time processing on both GPUs and CPUs.

% \keywords{Lookup Tables, Color Transforms, Real-time Image Processing, Image Enhancement}
\end{abstract}

%%%%%%%%% BODY TEXT
\section{Introduction}

% \noindent \textbf{brief review:} ISP and learnable 3D LUTs

% introduce traditional LUTs in ISP (1D \& 3D), briefly describe two different transforms (transforms in 3D LUTs can also cover those in 1D LUTs). high efficiency and flexible expressiveness promote recent advances in learnable LUTs: 1D or 3D solely

% \noindent \textbf{issues of a single 3D LUTs:}
% point out issues:
% 1D solely (ability to handle component-independent transforms, lack the precision required for more complex non-linear color correction such as film stock emulation and color grading)
% 3D solely (ability to handle both transforms in a coupled manner, trade-off of capacity and memory, insufficient cell utilization) (maybe provide some examples)
% simultaneously two transforms, coupled, input range maybe not well-defined, different LUT size is required
% LUT size is not trivial to adapt to different inputs, relatively large LUT size is required, heavy memory burden, insufficient cell utilization

The lookup table (LUT) is a promising data structure to efficiently conduct specific transforms by replacing expensive runtime computation with cheap array caching and indexing. It precomputes the outputs of a function over a sampled domain of inputs and evaluates the same function via efficient lookup and interpolation operations. LUTs are widely used to optimize color transforms in the image signal processor (ISP), a crucial component in the camera imaging pipeline and many display devices. To transform sensor signals into human-perceptible digital images, typical ISP follows the divide-and-conquer principle to utilize two types of LUTs, of 1D and 3D, for handling different transforms \cite{ECCV16software}. 1D LUTs are suitable for \textit{component-independent transforms} that require no interaction between three color components, such as white-balancing, gamma correction, brightness adjustment, and contrast stretching. 3D LUTs further enable the mixture of different color components, thus supporting more sophisticated \textit{component-correlated transforms}, like adjustments in hue and saturation.

The high effectiveness and efficiency of LUTs motivate recent advances in deep learning to propose learnable, image-adaptive LUTs for enhanced real-time image enhancement \cite{ICPR21curl,ICCV21Star,ICCV21RCT,ECCV20Global,Arxiv20flexible,CVPR20zero,TOG18exposure,CVPR18DisRec,TIP20personalized,TPAMI203DLUT,ICCV21SALUT}. % However, these methods encode the complete capability of color transform to only a single type of LUTs, either 1D or 3D, not considering the conjunction of both types, which makes them suffer from respective inner drawbacks, thus limiting their expressiveness. Specifically, methods based on 1D LUTs lack the critical model capability of interacting component information as they transform each color component independently.
However, these methods encode a complete color transform to only a single type of LUTs, either 1D or 3D, but neglecting the limited capability of a single module to model component-independent and component-correlated transformations simultaneously. Such a paradigm limits the expressiveness of these methods. Specifically, methods based on 1D LUTs lack the critical model capability of interacting component information as they work on each color component independently. 
Though the methods based on 3D LUTs are able to handle both component-independent and component-correlated transforms, they model these two transforms in a coupled manner, which increases the capability requirement of the model. The reason owns to the lack of a prior component-independent transform that can rescale the input image range into a normalized and perceptually uniform color space for the 3D LUTs. Therefore, the 3D LUTs rely on increasing their sizes for adaption to the diversity of input color ranges. For example, \cite{TPAMI203DLUT,ICCV21SALUT} adopt 33-point 3D LUTs, while the ISP typically employs 17-point or even 9-point 3D LUTs \cite{ECCV16software}. The large LUT size introduces massive parameters, resulting in heavy memory burden and high training difficulty. Besides, adopting a relatively large LUT size will lead to insufficient cell utilization of the 3D LUTs since the colors appearing in a single input image usually occupy only a tiny sub-space of the entire color space, causing redundancy of the model capacity.

% \noindent \textbf{solution:}
% inspired by the divide-and-conquer insight of ISP, decouple a single color transform into component-independent and component-correlated transform, instantiated as 1D LUTs and 3D LUTs.
% \noindent \textbf{advantage:}
% (bidirectional):
% 3D to 1D: compensate the ability to handle color interaction (mixup different color channels)
% 1D to 3D: maps the input color space to a normalized, perceptually uniform color space.
% improve cell utilization, reduce redundancy
% reduce the precision requirement of the 3D LUTs, good nature eases model quantization, lightweight

\begin{figure}[t]
    \centering
    \includegraphics[width=\linewidth]{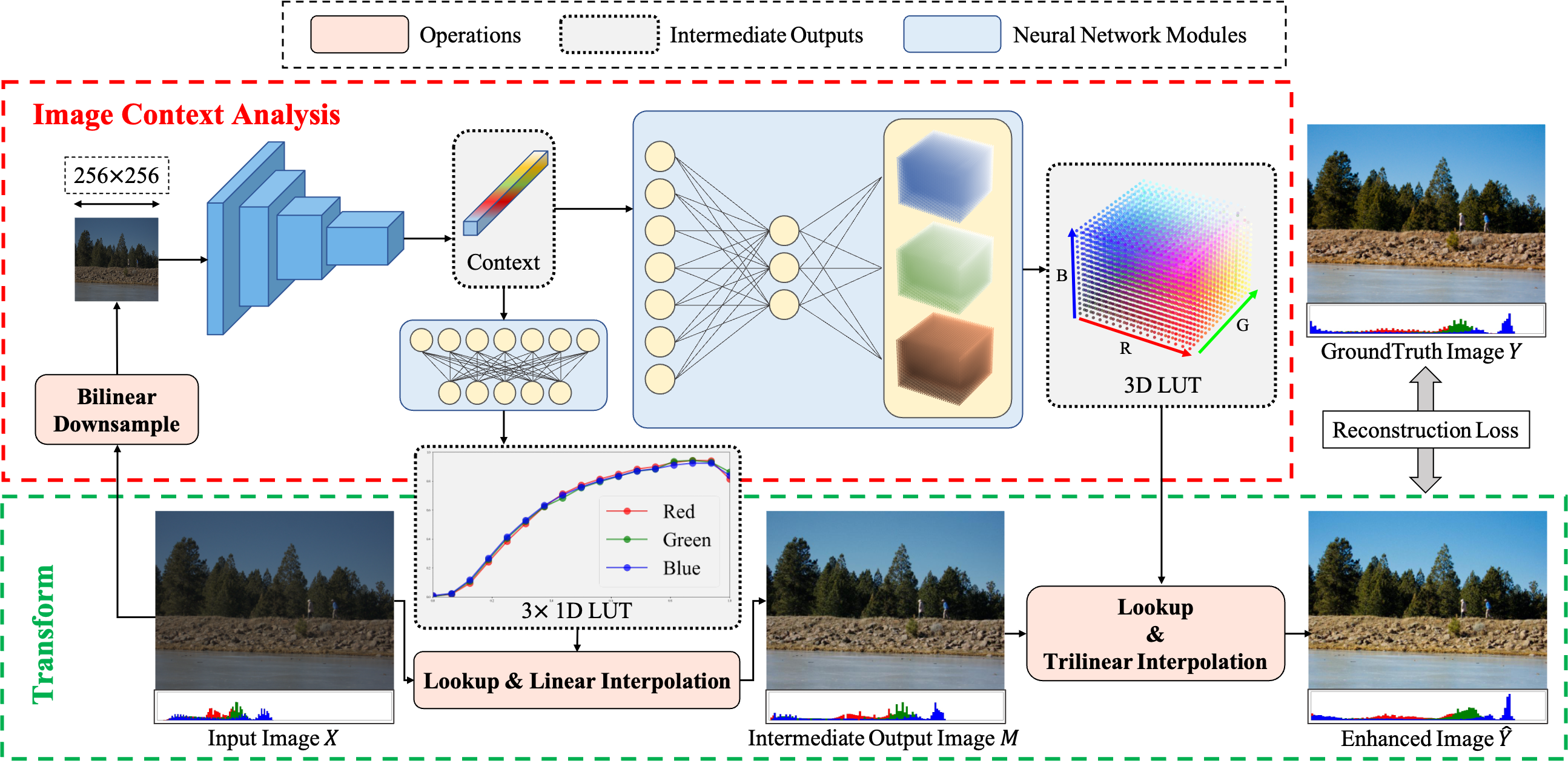}
    \caption{Framework of the proposed method. Our method employs a lightweight CNN network to analyze the image context from a down-sampled, fixed-resolution version of the input image. The image context is used to guide the generation of a $3\times$ 1D LUT and a 3D LUT in an image-adaptive fashion. The original input image is afterward enhanced by the cascade of the predicted 1D and 3D LUTs. Best viewed in color.}
    \label{fig:framework}
\end{figure}

To simultaneously improve the model's expressiveness and efficiency, we propose a novel framework called separable image-adaptive lookup tables (SepLUT). It decouples a single color transform into component-independent and component-correlated sub-transforms instantiated as 1D and 3D LUTs, respectively. The idea is directly motivated by the common practice in ISP, where 1D and 3D LUTs play their roles in conjunction. As illustrated in \Cref{fig:framework}, we follow the paradigm of dynamic neural functions \cite{NIPS16DFN} to employ a CNN backbone network on a downsampled, fixed-resolution version of the input image for predicting the parameters of a 3$\times$ 1D LUT and a 3D LUT. The two generated LUTs are then applied to the original input image sequentially -- the 3$\times$ 1D LUT rescales each color channel to adaptively adjust the brightness/contrast, followed by the 3D LUT to mix up three color components for manipulation on hue and saturation. The advantages are two-fold. On the one hand, the 3D LUTs can complement the 1D LUTs with color components interaction. On the other, the 1D LUTs can redistribute the input colors into specific ranges of the following 3D LUTs, which increase the cell utilization of the 3D LUTs, thus reducing the redundant capacity and enabling the usage of smaller sizes. Furthermore, the consistency between the input and output spaces of a LUT allows trivial LUT quantization that provides significant lightweight property to the proposed method. 

The contributions of this paper are three-fold: (1) We present a novel viewpoint of separating a single color transform into two sub-transforms, \textit{i.e.}, the component-independent transform and the component-correlated transform. (2) We propose a general framework that adopts a cascade of 1D and 3D LUTs to instantiate the above two sub-transforms, making them facilitate each other and present overall lightweight property, high efficiency, and enhanced expressiveness. (3) We demonstrate the efficiency and effectiveness of the proposed method via extensive experiments on publicly available benchmark datasets.

\section{Related Works}

\subsection{Lookup Tables}

A lookup table (LUT) defines a table of values addressed by a set of indices. It usually serves as an effective and efficient representation of a univariate or multivariate function $y=f(x_1, \dots, x_n), n=1,2,\dots$ by enumerating all possible input combinations $\{(x_1,\dots,x_n)\}$ and storing the corresponding output values $y$. The function can afterward be evaluated using only the memory access and interpolation, without performing the computation again. Therefore, LUTs are commonly used in computer systems \cite{kwok1995efficient,sharif2014high}, especially some embedded devices \cite{ECCV16software}, to accelerate computation. The simplest LUT is the 1D LUT indexed by a single variable ($n=1$), which utilizes linear interpolation to generate values of specific indices that are not an element of the table. Another frequently-used LUTs are the 3D LUTs indexed by a triplet of three independent variables ($n=3$), which require more complicated interpolation techniques such as the trilinear \cite{selan2005using} and the tetrahedral \cite{kasson1995performing} interpolations. According to \cite{ECCV16software}, most of the modules in the practical ISP systems are implemented using either 1D or 3D LUTs due to their high efficacy and suitability for modeling color transforms.

\subsubsection{Learnable LUTs}
The high efficiency and wide usage of the LUTs in ISP also attract efforts in deep learning-based image enhancement to learning more powerful LUTs via data-driven approaches. Previous works \cite{ICPR21curl,ICCV21Star,ICCV21RCT,ECCV20Global,Arxiv20flexible,CVPR20zero,TOG18exposure,CVPR18DisRec,TIP20personalized} mainly focus on learning 1D LUTs to mimic the color adjustment curves in popular image enhancement software such as Photoshop and Lightroom. They regress either a set of the control points of the curve, or the coefficients of some hand-crafted functions (\textit{e.g.}, polynomial function). However, these methods usually suffer from the lack of correlation between color channels. Recently, \cite{TPAMI203DLUT,ICCV21SALUT} extended those 1D LUT-based methods into using 3D LUTs. They predict 3D LUTs with adaption to different image contents by learning several image-independent basis 3D LUTs and combining them using image-dependent weights. The 3D LUTs consider the relationship between color channels and thus provide higher expressiveness to model more complicated color transforms. However, the large model capacity of the 3D LUTs requires massive learnable parameters, making these methods suffer from heavy memory/storage footprints. In contrast to previous works based on a single kind of LUTs, we present a more general framework that considers the conjunction of both types of LUTs, thus providing an advanced solution to the above limitations.

\subsection{Image Enhancement}

Existing learning-based image enhancement methods can be roughly categorized into two paradigms, \textit{i.e.}, the fully convolutional network-based methods, and the color transform-based methods.  The first paradigm \cite{CVPR18DPE,CVPR20DeepLPF,ACMMM18AD,ACMMM19Lowlight,BMVC18Lowlight,CVPR18Dark} is to train a fully convolutional network (FCN) that directly regresses the enhanced image from the input in a dense prediction fashion. However, these methods are still far from practical applications due to their heavy computational burdens and limited feasible input resolutions. In contrast, the second paradigm decouples the color transformations from the heavy CNN model for real-time and high-resolution processing. Specifically, these methods employ CNNs on a low-resolution, fixed-size version of the input image to predict image-adaptive parameters of some specific color transform functions. Typical color transform functions include affine transformation matrices \cite{TOG17HDRNet,CVPR19UPE,WACV20supervised,ACCV20color}, curve-based functions \cite{ICPR21curl,ICCV21Star,ICCV21RCT,ECCV20Global,Arxiv20flexible,CVPR20zero,TOG18exposure,CVPR18DisRec,TIP20personalized}, multi-layer perceptrons (MLPs) \cite{ECCV20CSRNet} and 3D LUTs \cite{TPAMI203DLUT,ICCV21SALUT}. These learned transform functions can adapt to different input image contents and present high runtime efficiency. Some of them are also flexible and scalable on arbitrary input resolutions. Therefore, our work also follows this line of the color transform-based scheme and builds a novel framework with a cascade of 1D and 3D LUTs for real-time image enhancement.
\section{Methods}

\begin{table}[t]
    \begin{minipage}{0.52\linewidth}
        \centering
        \caption{Architecture of the backbone network, where $m$ is a hyper-parameter that serves as a channel multiplier controlling the width of each convolutional layer.}
        \begin{tabular}{ccc}
            \toprule[1pt]
            Id & Layer           & Output Shape \\
            \midrule[1pt]
            0  & Bilinear Resize    & $3\times 256\times 256$ \\
            1  & Conv3x3, LeakyReLU & $m\times 128\times 128$ \\
            2  & InstanceNorm       & $m\times 128\times 128$ \\
            3  & Conv3x3, LeakyReLU & $2m\times 64\times 64$ \\
            4  & InstanceNorm       & $2m\times 64\times 64$ \\
            5  & Conv3x3, LeakyReLU & $4m\times 32\times 32$ \\
            6  & InstanceNorm       & $4m\times 32\times 32$ \\
            7  & Conv3x3, LeakyReLU & $8m\times 16\times 16$ \\
            8  & InstanceNorm       & $8m\times 16\times 16$ \\
            9  & Conv3x3, LeakyReLU & $8m\times 8\times 8$ \\
            10 & Dropout (0.5)      & $8m\times 8\times 8$ \\
            11 & AveragePooling     & $8m\times 2\times 2$ \\
            12 & Reshape            & $32m$ \\
            \bottomrule[1pt]
        \end{tabular}
        \label{tab:arch-backbone}
    \end{minipage}
    \hspace{1pt}
    \begin{minipage}{0.48\linewidth}
        \centering
        \caption{Architecture of the 1D LUT generator, where $S_o$ is the size (number of elements) of the 1D LUTs. "FC" denotes the fully-connected layer.}
        \begin{tabular}{ccc}
            \toprule[1pt]
            Id & Layer & Output Shape \\
            \midrule[1pt]
            0 & FC & $3S_o$ \\
            1 & Reshape & $3\times S_o$ \\
            2 & Sigmoid & $3\times S_o$ \\
            \bottomrule[1pt]
        \end{tabular}
        \label{tab:arch-1dlut-generator}
        \vspace{5pt}
        \centering
        \caption{Architecture of the 3D LUT generator, where $S_t$ is the size (number of elements along each dimension) of the 3D LUT.}
        \begin{tabular}{ccc}
            \toprule[1pt]
            Id & Layer & Output Shape \\
            \midrule[1pt]
            0 & FC & $K$ \\
            1 & FC & $3S_t^3$ \\
            2 & Reshape & $3\times S_t\times S_t\times S_t$ \\
            \bottomrule[1pt]
        \end{tabular}
        \label{tab:arch-3dlut-generator}
    \end{minipage}
\end{table}

\subsection{Overall Framework}

% objectives: real-time and lightweight
% dynamic neural functions
% a backbone network to predict image-adaptive color transform function
% function is decoupled
% apply decoupled transforms sequentially
% we show that the method has good properties for cuda speedup and quantization

\Cref{fig:framework} shows an overview of the proposed image enhancement framework, which follows the popular paradigm of dynamic neural functions. Specifically, a lightweight CNN network is employed as the backbone network on the input image to extract global context that will serve as the guide to generate image content-dependent color transform functions. The form of the color transform functions is designed as a cascade of a 3$\times$ 1D LUT and a 3D LUT, aiming to handle the component-independent and component-correlated transforms in a decoupled and sequential manner. Afterward, the generated functions enhance the quality of the input through efficient lookup and interpolation operations.

\subsection{Global Image Context Analysis: Backbone Network}

The backbone CNN network is central to achieving image-adaptiveness in the proposed framework. 
It is responsible for predicting the parameters of the subsequent 1D and 3D LUTs according to the image content through analyzing the input image $X\in[0, 1]^{3\times H\times W}$.
% It is responsible for analyzing the input image $X\in[0, 1]^{3\times H\times W}$ and predicting the parameters of the subsequent 1D and 3D LUTs according to the image content.
Since both the 1D and 3D LUTs in our method are designed for global color transform, the backbone network only needs to capture a coarse understanding of the input image. Therefore, a low-resolution version (\textit{e.g.}, $256\times 256$) of the input image is sufficient and can substantially reduce the computational cost to a fixed level. The detailed architecture of the CNN backbone network is listed in \Cref{tab:arch-backbone}, where $m$ is a hyper-parameter controlling the channel width of each convolutional layer. The backbone network adopts 5 strided convolutional layers to downsample the input image into $1/32$ resolution. % Each convolutional layer is followed by a leaky ReLU activation and an instance normalization.
At the end of the network are an average pooling layer and a reshape operation that further convert the feature maps into a compact vector representation $E\in\mathbb{R}^{32m}$. The vector representation captures some global attributes of the input image and will be fed into the subsequent modules as the guide to generate image content-dependent LUT parameters.

\subsection{Component-independent Transform: 1D Lookup Tables}
\label{sec:method-1d}

The component-independent transform aims to redistribute the input colors into a more perceptually uniform space that can increase the cell utilization of the following 3D LUT. In this paper, we propose to adopt the $3\times$ 1D LUT for the above purpose, where three individual 1D LUTs are predicted for each color channel, respectively.
The elements $\{T^c_{1D}\}_{c\in\{r,g,b\}}$ of the 1D LUTs are conditioned on the image context $E$ to achieve image-adaptiveness, formulated as:
\begin{equation}
    \{T^r_{1D}, T^g_{1D}, T^b_{1D}\} = g_{1D}(E),
\end{equation}
where $T^c_{1D}\in[0,1]^{S_o}$ denotes a 1D LUT of $S_o$ values for the channel $c\in\{r,g,b\}$. $g_{1D}$ is the 1D LUT generator module that takes $E$ as input and predicts all output values, whose architecture is detailed in \Cref{tab:arch-1dlut-generator}. Note that the \textit{sigmoid} layer is used to normalize the elements in the 1D LUTs into a valid range. Once the 1D LUTs are predicted, the component-independent transform can be performed on each pixel separably via simple linear interpolation:
\begin{equation}
    \label{eq:1d-transform}
    M[c,h,w] = \text{linear\_interpolate}(T^c_{1D}, X[c,h,w]),
\end{equation}
where $M\in[0,1]^{3\times H\times W}$ denotes the intermediate image transformed by the 1D LUTs. $h\in\mathbb{I}_0^{H-1}$ and $w\in\mathbb{I}_0^{W-1}$ are indices to traverse the image \footnote{$\mathbb{I}_s^t$ denotes the integer set starting from $s$ and ending at $t$, \textit{i.e.}, $\mathbb{I}_s^t=\{s, \dots, t\}$.}. 

Intuitively, directly applying histogram equalization, a conventional technique in image processing to adjust image contrast, is another alternative as the component-independent transform to increase the distribution uniformity of each color component. However, histogram equalization maximizes the entropy of the image statically by mapping the input color ranges to an \textit{exact uniform} distribution, which is not intelligent and not always necessary in the scenario of image enhancement. Instead, by adopting the data-driven approach to learn the 1D LUTs, our method is expected to be image-adaptive for the network and each input image. The quantitative comparisons can be found in \Cref{tab:exp-abla-decoupling}.

\subsection{Component-correlated Transform: 3D Lookup Tables}
\label{sec:method-3d}

After the previous component-independent transform, the component-correlated transform aims at mixing and interacting different color channels to achieve more sophisticated color transforms, such as alteration in hue and saturation. Considering the balance between expressiveness and efficiency, we choose the 3D LUTs to formulate such a transformation from a triplet to one another. A typical 3D LUT defines a 3D grid of $S_t^3$ elements, where $S_t$ denotes the number of values along each color dimension. Similar to \Cref{sec:method-1d}, all the $S_t^3$ elements in the 3D LUT should be automatically predicted by the neural network to consider the adaption to the diversity of various input images. Such an objective formulates a mapping from the image context $E$ to a $3S_t^3$-dimension parameter space:
\begin{equation}
    \label{eq:3d_generator}
    T_{3D} = g_{3D}(E),
\end{equation}
where $T_{3D}\in[0,1]^{3\times S_t\times S_t\times S_t}$ denotes a 3D LUT of size $S_t$. $g_{3D}$ is the 3D LUT generator module. To prevent the involvement of significant memory burden and training difficulty, we consider the \textit{rank factorization} to decompose the complete mapping in \Cref{eq:3d_generator} into two sub-mappings $h_1$ and $h_2$:
\begin{equation}
    \label{eq:rank_factorization}
    g_{3D} =\ h_1 \circ h_2,\ h_1:\mathbb{R}^{32m}\rightarrow\mathbb{R}^K,\ h_2:\mathbb{R}^K\rightarrow\mathbb{R}^{3S_t^3}.
\end{equation}
$h_1$ and $h_2$ can be instantiated with two respective fully-connected (FC) layers. Compared with a single FC layer, such a strategy reduces the number of parameters from $32m\times 3S_t^3$ to $K\times(32m+3S_t^3)$, making the transform more feasible and easier to optimize. The detailed architecture can be found in \Cref{tab:arch-3dlut-generator}.

Given the predicted 3D LUT $T_{3D}$, the final enhanced image $\hat{Y}\in[0,1]^{3\times H\times W}$ can be derived via a simple trilinear interpolation:
\begin{equation}
    \label{eq:3d-transform}
    \hat{Y}[c,h,w] = \text{trilinear\_interpolate}(T_{3D}, M[c,h,w]).
\end{equation}

\subsection{Efficient Implementation via Quantization}

% \subsubsection{Kernel Merging}
% Since the transforms performed by both the 1D LUTs (\Cref{eq:1d-transform}) and the 3D LUT (\Cref{eq:3d-transform}) are independent between different pixels, they can be substantially accelerated by parallelizing them via customized CUDA codes. Besides, thanks to the non-overlapping of the dependencies of \Cref{eq:1d-transform} and \Cref{eq:3d-transform}, they can be further merged into a single CUDA kernel to prevent duplicate memory access during multiple image pixel traversals. Therefore, the proposed method presents high efficiency. 

% \subsubsection{Quantization on Lookup Tables}
\label{sec:method-quan}
Our method can also benefit from the model quantization technique to further reduce the memory and storage footprints. Specifically, the FC layers in the LUT generators (\Cref{tab:arch-1dlut-generator,tab:arch-3dlut-generator}) are equivalent to learning several image-independent basis LUTs encoded as the learnable parameters. The input to the FC layer serves as the image-dependent coefficients that linearly combine the basis LUTs into the final LUT.
Note that the output of the LUT is simply a linear combination of the elements in the LUT and falls into the color space that can be naturally quantized.
Therefore, thanks to the semantic consistency between the parameter space and the output space, the trained parameters of the LUT generators can be naturally quantized to lower-bit representation during the testing time without significant performance decline. It is worth noting that it is not trivial for other image enhancement methods to benefit from the model quantization techniques due to their inconsistency between parameter and output spaces. More complicated model quantization approaches are required but would introduce cumbersome training protocols or substantially hurt the performance. Besides, since the LUTs and the input image can be quantized into fixed-point representation (\textit{e.g.}, 8-bit integer), we can also replace the floating-point computation with the fixed-point counterpart in the lookup and interpolation procedure for further speedup, as shown in \Cref{tab:exp-quantization}. 

% \subsection{Loss Function}

% The proposed framework can be trained in an end-to-end manner. The loss function is simply the mean square error (MSE) loss. We do not introduce any other constraint or loss function to the predicted 1D and 3D LUTs, willing that they can be image-adaptive for the network and input image itself, not for any hand-crafted priors.
\section{Experiments}
\label{sec:exp}

\subsection{Datasets}

The publicly available MIT-Adobe FiveK \cite{CVPR11FiveK} and PPR10K \cite{ICCV21PPR10K} datasets are adopted to evaluate the proposed method. The FiveK dataset contains 5,000 RAW images with five manually retouched ground truths (A/B/C/D/E). Version C is selected in our experiments. We use the commonly used settings \cite{CVPR18DPE,TPAMI203DLUT} to split the dataset into 4,500 image pairs for training and the remaining 500 image pairs for testing. The 480P version of the dataset is used to speed up the training process, while the testing is conducted on both 480P and original 4K resolutions. The PPR10K dataset contains a larger scale of 11,161 RAW portrait photos with 3 versions of groundtruths (a/b/c). We follow the setting in \cite{ICCV21PPR10K} to utilize all three retouched versions as the groundtruth in different experiments and split the dataset into 8,875 pairs for training and 2,286 pairs for testing. Experiments are conducted on the 360P verison of the dataset.
Following \cite{TPAMI203DLUT}, experiments are organized on two typical application scenarios, \textit{photo retouching}, and \textit{tone mapping}. The former task retouches the input image in the same sRGB format, whereas the latter transforms the 16-bit CIE XYZ input images into 8-bit sRGB. We conduct both tasks on the FiveK dataset, but only the retouching task on PPR10K as done in \cite{ICCV21PPR10K}.
As for the data augmentation strategies, we follow the settings in \cite{TPAMI203DLUT} and \cite{ICCV21PPR10K} to ensure a fair comparison. 

\subsection{Implementation Details}
\Cref{tab:arch-backbone,tab:arch-3dlut-generator,tab:arch-1dlut-generator} show the instantiations of modules in the proposed method. The parameters of the backbone network are randomly initialized as in \cite{Xavier}. To make a fair comparison, we also conduct experiments using the ResNet-18 \cite{ResNet} backbone network (initialized with ImageNet-pretrained \cite{ImageNet} weights) on the PPR10K dataset, as done in \cite{ICCV21PPR10K}.
As for the LUT generators, inspired by \cite{TPAMI203DLUT}, we initialize the 3D LUT generator to predict an identity mapping at the early training stage to speed up the training convergence.
The mean square error (MSE) loss is adopted to train the proposed method in an end-to-end manner. We do not introduce any other constraint or loss function to the predicted 1D and 3D LUTs, willing that they can be image-adaptive for the network and input image itself, not for any hand-crafted priors. We implement our method based on PyTorch 1.8.1 \cite{PyTorch}. The standard Adam optimizer \cite{Adam} is adopted to train the proposed method, with the mini-batch size set to 1 and 16 on FiveK and PPR10K, respectively.
All models are trained for 400 epochs with a fixed learning rate of $1\times 10^{-4}$ on an NVIDIA Tesla V100 GPU. $K$ in \Cref{eq:rank_factorization} is set to 3 and 5 for FiveK and PPR10K respectively, whereas $S_o$ and $S_t$ are set according to the purposes of the experiments. We provide them in the following sections. % The source code and pre-trained models are publicly released to promote research reproducibility.

\subsection{Ablation Studies}

In this section, we conduct several ablation studies on the retouching task with the FiveK dataset (480P) to verify the key components of the proposed method.

\subsubsection{Size of Lookup Tables}

\begin{figure}[t]
    \centering
    \subfloat[Different sizes ($S_o$) of the 1D LUT.]{\centering\includegraphics[width=0.49\linewidth]{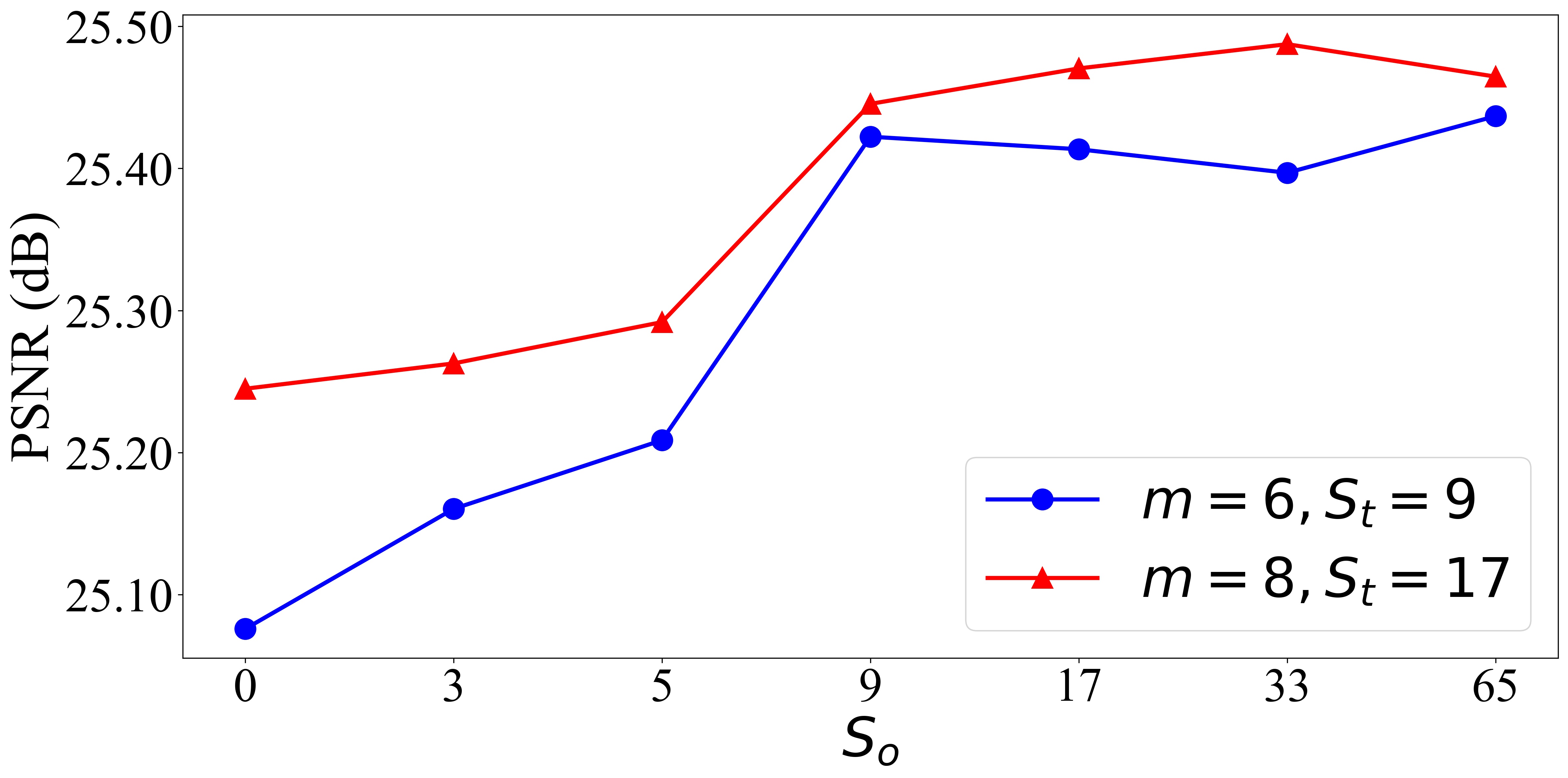}\label{fig:exp-abla-1dsize}}
    \subfloat[Different widths ($m$) of the backbone.]{\centering\includegraphics[width=0.49\linewidth]{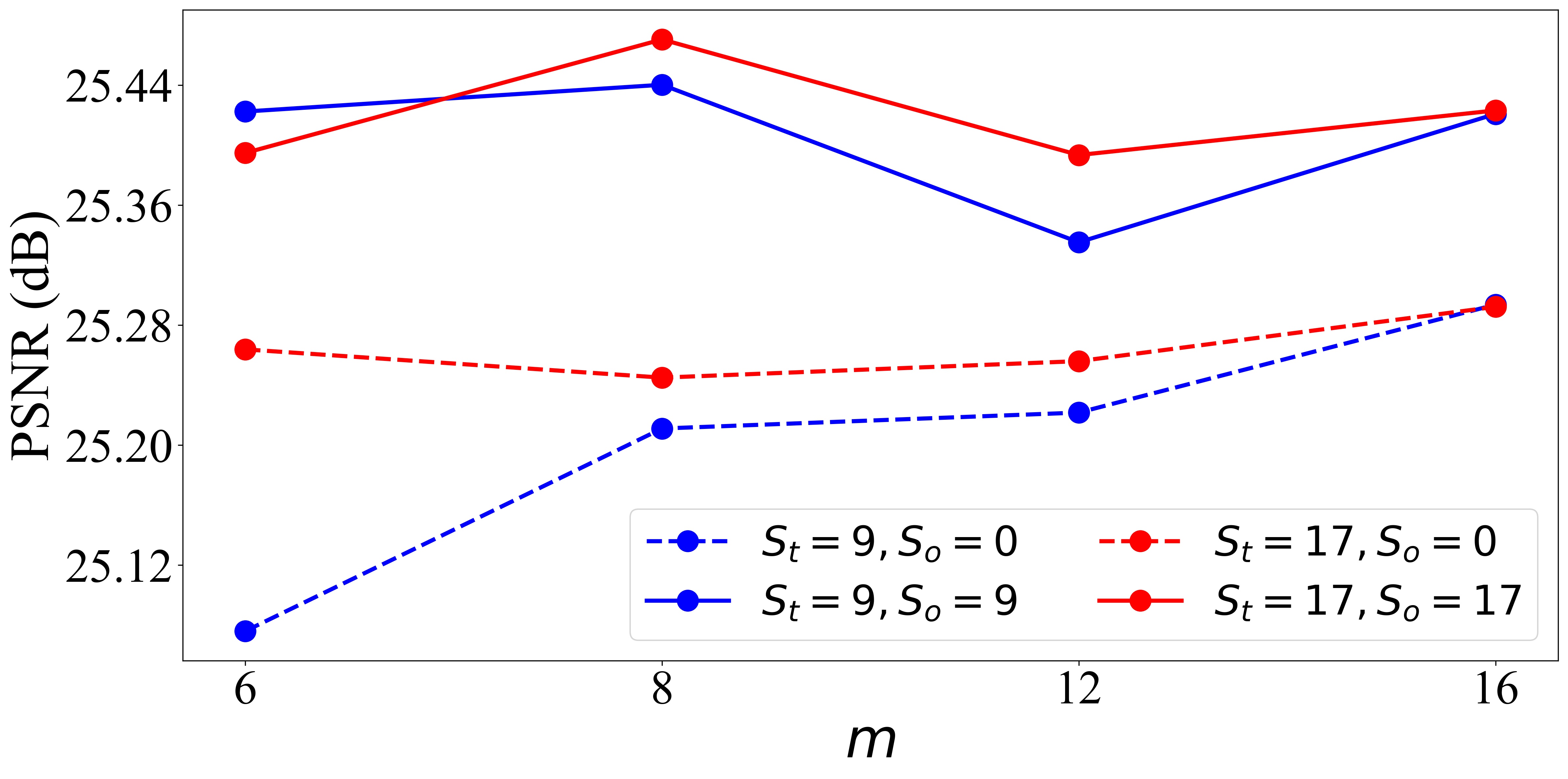}\label{fig:exp-abla-backbone}}
    \caption{Ablation studies on different sizes of the 1D LUT (a) and different widths of the backbone network (b). $S_o=0$ indicates models with only 3D LUTs.}
    \label{fig:exp-abla-1dsize-backbone}
\end{figure}

(1) \textbf{3D LUT}: We explore the effects of the size of the 3D LUT by varying $S_t$ in the absence of the 1D LUT ($S_o=0$) and $m=8$. The experiments on the FiveK dataset (480P) for the photo retouching task show that, with only the 3D LUT, decreasing the LUT size $S_t$ (from 33, 17 to 9) can significantly reduce the number of parameters (from 385K, 106K to 69K) without a substantial performance drop (from 25.27dB, 25.24dB to 25.21dB). Such a phenomenon suggests the capacity redundancy of the 3D LUT. Therefore, considering the balance between performance and model size, we select $S_t=9$ and $S_t=17$ as the default settings in this paper.
(2) \textbf{1D LUT}: We also investigate the effects of different sizes of the 1D LUT when the size of the 3D LUT is fixed. As shown in \Cref{fig:exp-abla-1dsize}, increasing the size of the 1D LUT improves the performance continuously and saturates after it exceeds that of the 3D LUT. A possible reason is that the precision of the 3D LUT serves as a bottleneck and will cancel the additional quantization granularity introduced by the 1D LUT.

\subsubsection{Capacity of the Backbone}
Since the backbone network is responsible for providing a coarse analysis of the input image to guide the generation of the 1D and 3D LUTs, its capacity requirement should be correlated with the capacity or the size of the LUTs. For verification, we ablate the width of the backbone network by varying the hyper-parameter $m$ under the same setting of the LUT sizes ($S_o$ and $S_t$) and report the results in \Cref{fig:exp-abla-backbone}. The ablation results indicate that increasing the width of the backbone does not guarantee enhanced performance but might increase the capacity redundancy and the training difficulty. Besides, a larger LUT size is inclined to require a stronger backbone network. Considering the trade-off between the performance and the memory footprint, we adopt $m=6$ and $m=8$ for $S_t=9$ and $S_t=17$, respectively.

\begin{table}[t]
    \centering
    \caption{Ablation study on different instantiations of the component-independent transform. The results on the FiveK dataset (480P) for the photo retouching task are listed. "HE" is the abbreviation of Histogram Equalization. The $\uparrow$ and $\downarrow$ symbols indicate the larger or the smaller is better, respectively.}
    \setlength{\tabcolsep}{4pt}
    \begin{tabular}{ccccccc}
    \toprule[1pt]
    \multirow{2}{*}{Strategies} & \multicolumn{3}{c}{$m=6,S_t=S_o=9$} & \multicolumn{3}{c}{$m=8,S_t=S_o=17$} \\
    \cline{2-7}
     & PSNR $\uparrow$ & SSIM $\uparrow$ & \#Params $\downarrow$ & PSNR $\uparrow$ & SSIM $\uparrow$ & \#Params $\downarrow$ \\
    \midrule[1pt]
    HE & 21.75 & 0.848 & 42.0K & 21.76 & 0.847 & 106.7K \\
    1D LUT & 25.32 & 0.918 & 43.7K & 25.41 & 0.917 & 111.1K \\
    3$\times$ 1D LUT & 25.42 & 0.921 & 47.2K & 25.47 & 0.921 & 119.8K \\
    \bottomrule[1pt]
    \end{tabular}
    \label{tab:exp-abla-decoupling}
\end{table}

\begin{table}[t]
    \centering
    \caption{Effects of the quantization technique. The runtimes are measured on 480P input using an Intel(R) Xeon(R) Platinum 8163 CPU, whereas the memory footprints are represented by the number of the equivalent parameters.}
    \setlength{\tabcolsep}{4pt}
    \begin{tabular}{cccccccc}
        \toprule[1pt]
        Method & \multicolumn{2}{c}{PSNR$\uparrow$} & \multicolumn{2}{c}{CPU Runtime $\downarrow$} & \multicolumn{3}{c}{\#Params$\downarrow$} \\
        \cline{2-8}
         & ori. & quant. & ori. & quant. & ori. & quant. & rel. \\
        \midrule[1pt]
        3D-LUT \cite{TPAMI203DLUT} & 25.28 & 25.25 & 17.35 & 15.52 & 593.5K & 332.5K & 43.98\%$\downarrow$\\
        $m=6$, $S_t=S_o=9$ & 25.42 & 25.35 & 25.34 & 15.65 & 47.2K & 37.9K & 19.59\%$\downarrow$ \\
        $m=8$, $S_t=S_o=17$ & 25.47 & 25.43 & 25.64 & 16.21 & 119.8K & 76.3K & 36.34\%$\downarrow$ \\
        \bottomrule[1pt]
    \end{tabular}
    \label{tab:exp-quantization}
\end{table}

\subsubsection{Instantiation of the Component-independent Transform}
% The purpose of the component-independent transform is to perform a prior adjustment that redistributes the input colors into a range that can reduce the capacity requirement of the following component-correlated transform.
In this section, we compare several variants to investigate the proper instantiation of the transform, including the histogram equalization (HE) transform, a learnable 1D LUT, and a learnable 3$\times$ 1D LUT. \Cref{tab:exp-abla-decoupling} demonstrates that the 3$\times$ 1D LUT performs the best under two different model settings. The results show that the fixed and hand-crafted mechanism of uniforming the input color distribution cannot adapt to different image contents and distinct retouching styles. Hence it does not guarantee advanced performance on the image enhancement task. The learnable 1D LUTs avoid the above issues by end-to-end optimization, and the 3$\times$ 1D LUT provides more flexibility and intelligence than the single 1D LUT.

\begin{figure}[t]
    \centering
    \includegraphics[width=\linewidth]{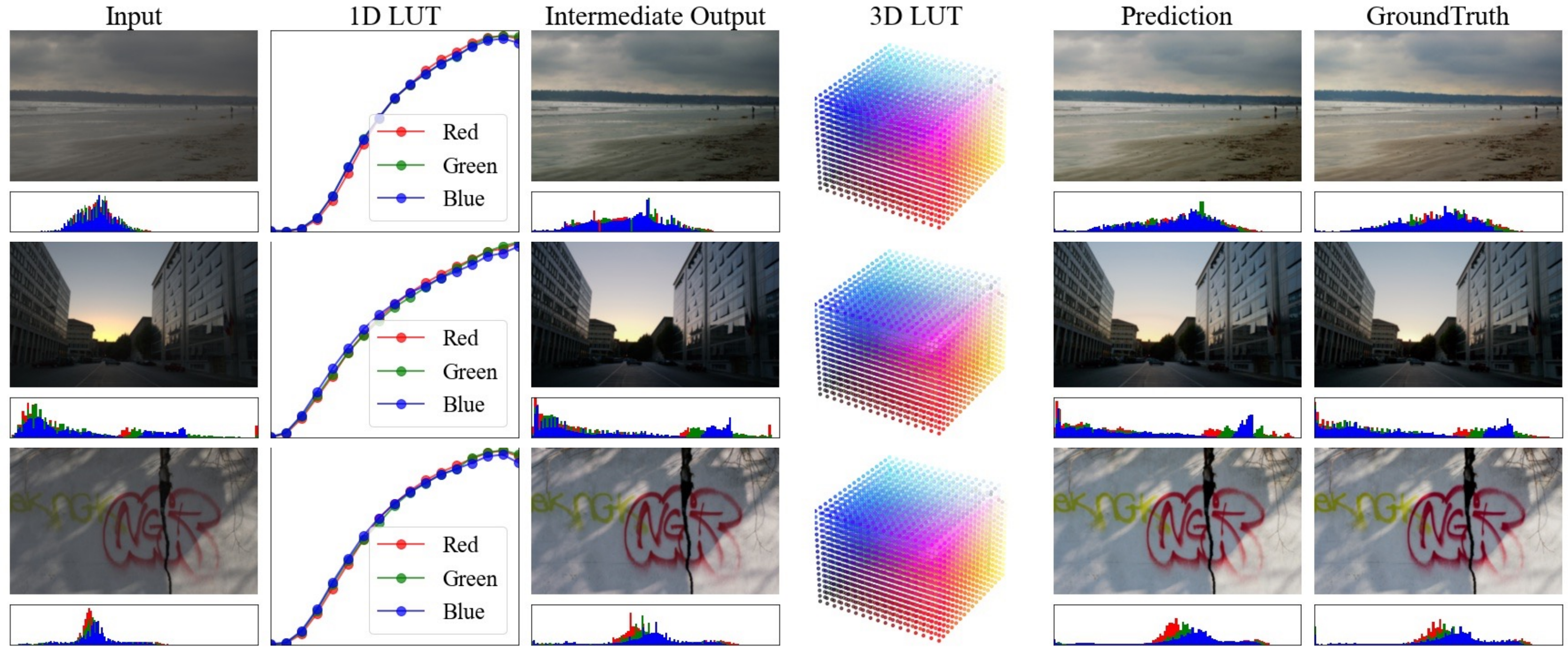}
    \caption{Illustration of the pair of images, the corresponding histograms, the predicted LUTs, the intermediate transformed output and the final prediction. Images are selected from the FiveK \cite{CVPR11FiveK} dataset (480P). Best viewed on screen.}
    \label{fig:procedure}
\end{figure}

\subsubsection{Quantization on Lookup Tables}
As described in \Cref{sec:method-quan}, we quantify the parameters in both 1D and 3D LUT generators from 32-bit floats to 8-bit integers. The results in \Cref{tab:exp-quantization} show that such quantization can significantly reduce the storage/memory footprints with only a slight performance drop. It is worth noting that the above results are obtained by directly quantizing the trained model without any finetuning or quantization-aware training strategy, demonstrating the flexibility of the proposed method. Besides, the LUT quantization enables the fixed-point arithmetic, which decreases the CPU inference time of our approach from about 25ms to 16ms. We also apply quantization and fixed-point arithmetic to another 3DLUT-based method \cite{TPAMI203DLUT} and find a similar phenomenon. The proposed framework benefits more from the fixed-point arithmetic in terms of runtime since both the 1D and 3D LUT transforms can be optimized, while \cite{TPAMI203DLUT} only includes the 3D LUT transform.

\subsection{Analysis}

To help draw an intuitive understanding of the behavior of the image-adaptive LUTs, we illustrate the intermediate outputs as shown in \Cref{fig:procedure}. It can be observed that the 1D LUT is inclined to adaptively stretch the input brightness and image contrast, making them in a state more similar to those of the ground truth. Afterward, the 3D LUT is responsible for altering the hue and enhancing the saturation. Besides, to provide quantitative analysis, we also calculate and analyze some statistics of a series of trained models, including the color distribution and the cell utilization of the 3D LUTs, as detailed in follows.

\begin{figure}[t]
    \centering
    \subfloat[$m=6, S_t=9$.]{\centering\includegraphics[width=0.49\linewidth]{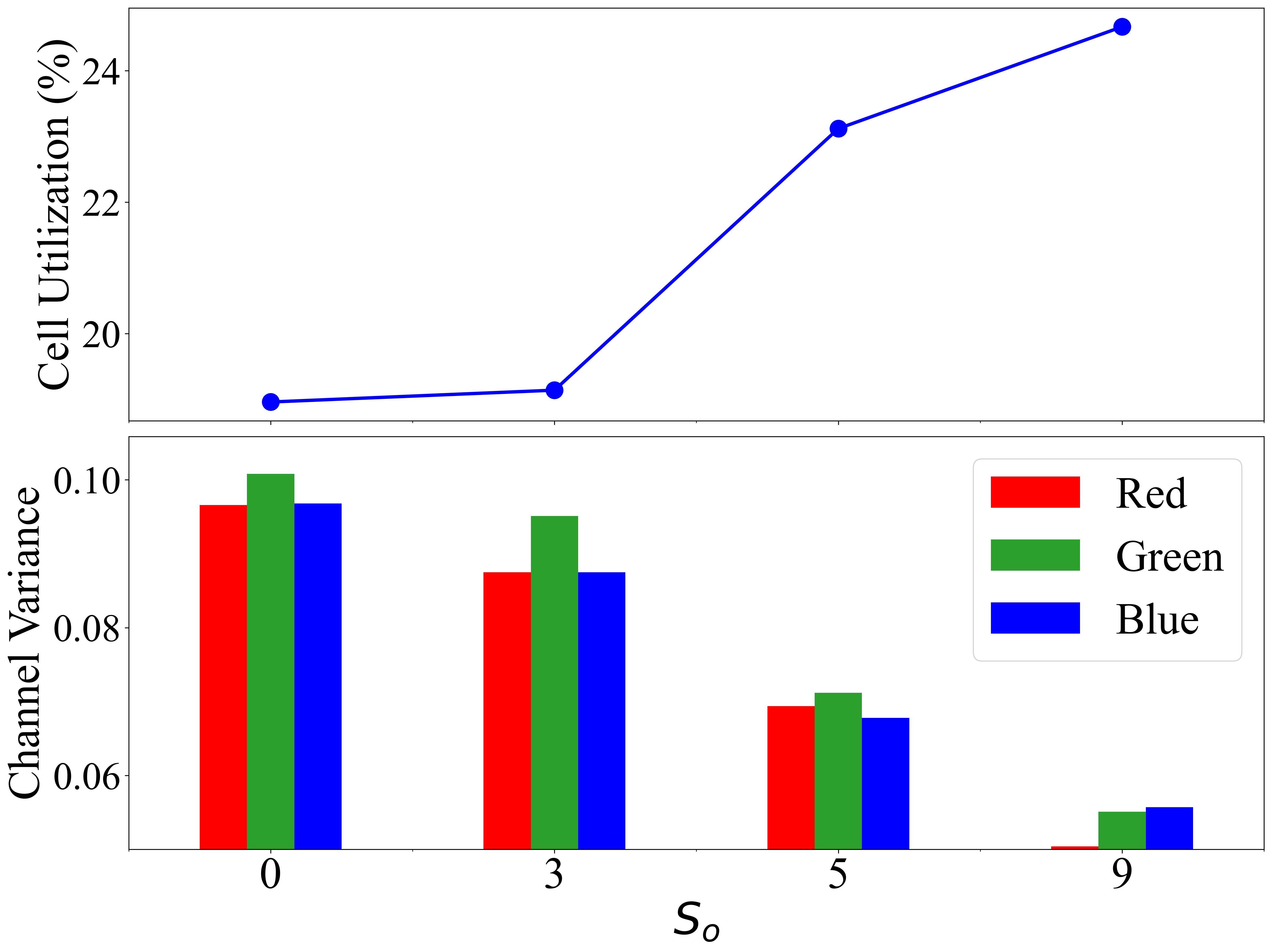}\label{fig:exp-analysis-m6v9}}
    \subfloat[$m=8, S_t=17$.]{\centering\includegraphics[width=0.49\linewidth]{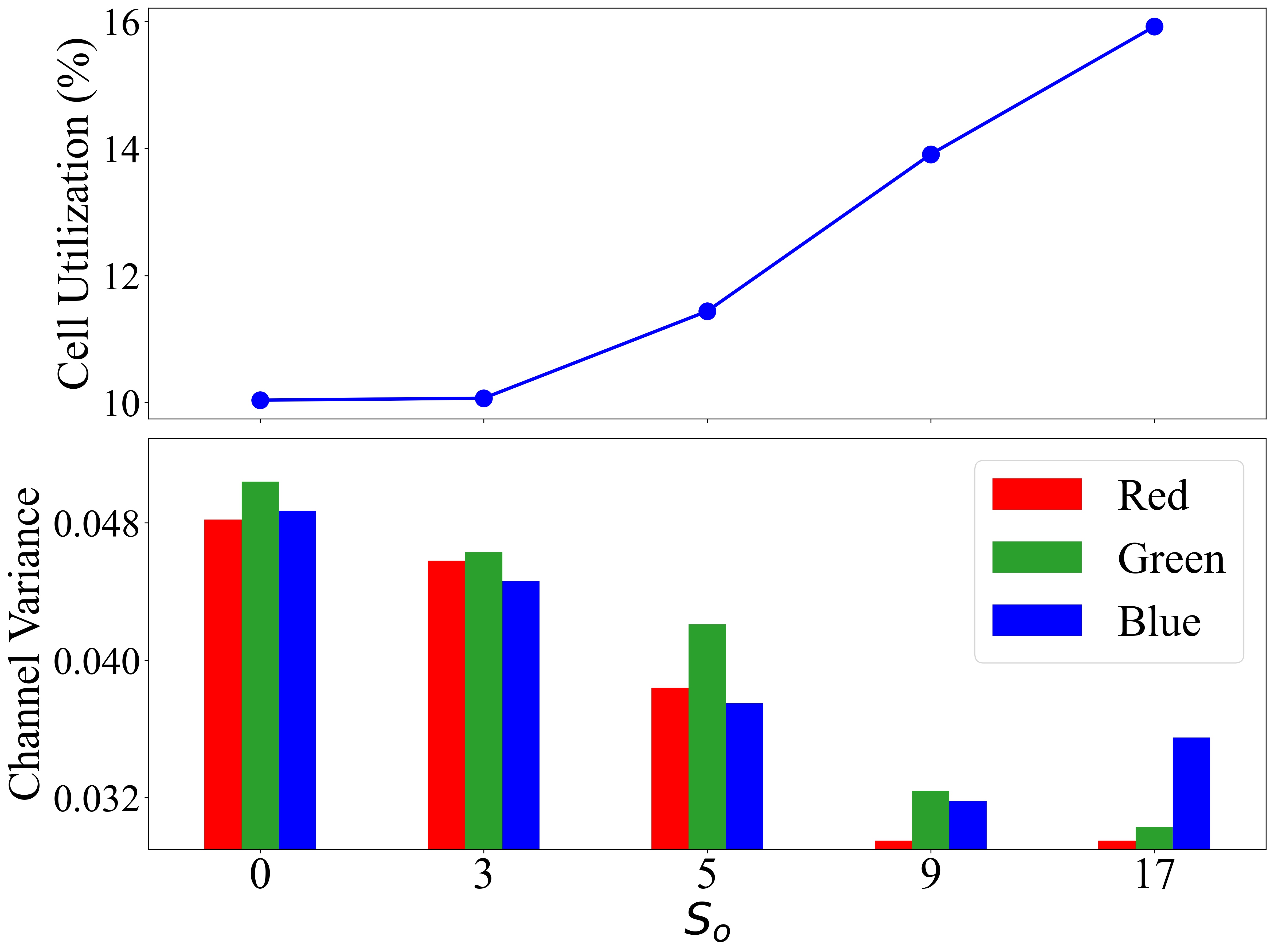}\label{fig:exp-analysis-m8v17}}
    \caption{Effects of the 1D LUT on the utilization of the 3D LUT (the top row) and the distribution of each color channel (the bottom row). Best viewed in color.}
    \label{fig:exp-analysis}
\end{figure}

\subsubsection{Distribution of Each Color Component}
To quantitatively investigate the effects of the 1D LUTs on the color distribution, we compare the histogram uniformity of images $M$ transformed by the 1D LUTs of different sizes. The histogram uniformity, to some extent, indicates the level of image contrast and can be approximated by the variance of the image histogram. The bottom row of \Cref{fig:exp-analysis} shows the results averaged on the FiveK dataset. As the size of the 1D LUT ($S_o$) increases, the per-channel histogram variance of the image decreases, % showing the uniformity of the color distribution of the input to the following 3D LUT is progressively increasing. This indeed increases the image contrast. However, an extremely uniform color distribution (after applied histogram equalization) would bring negative effects. The phenomenon is in line with our expectation that the 1D LUT would adjust the image brightness and contrast into a slightly uniform distribution in an image-adaptive manner.
showing the progressive increase of the uniformity of the color distribution. The phenomenon is in line with our expectation that the 1D LUT would adjust the image contrast into a more uniform distribution in an image-adaptive manner.

\subsubsection{Cell Utilization of 3D Lookup Tables}
A typical 3D LUT discretizes the entire 3D color space into a grid of cells. Unfortunately, when transforming a single input image, those cells are only partially utilized as the input rarely contains all possible colors. For example, the cell utilization of \cite{TPAMI203DLUT}, which can be regarded as a special case of our framework that adopts a single 33-point 3D LUT, is only about 5.53\%. To investigate the effects of the 1D LUT on the cell utilization of the 3D LUT, we count for each image the percentage of the cells that have valid pixels falling in. The results averaged on the FiveK dataset are reported in the top row of \Cref{fig:exp-analysis}. The 1D LUT is able to activate more cells to increase the model capability of the 3D LUT, and such ability becomes stronger as the 1D LUT becomes preciser (with larger $S_o$). % It is worth noting that the utilization improvement is far from saturation, inspiring future directions to further reduce the capacity waste for LUT-based image enhancement methods.

\subsection{Comparisons with State-of-the-Arts}

\begin{table}[t]
    \centering
    \caption{Quantitative comparisons on the \textbf{FiveK} dataset \cite{CVPR11FiveK} for \textbf{photo retouching}. "-" means the result is not available due to insufficient GPU memory. "*" indicates that the results are adopted from the original paper (some are absent ("/")) due to the unavailable source code. The best and second results are in \textcolor{red}{red} and \textcolor{blue}{blue}, respectively.}
    \setlength{\tabcolsep}{6pt}
    \begin{tabular}{cccccccc}
        \toprule[1pt]
        \multirow{2}{*}{Method} &
        \multirow{2}{*}{\#Params} & \multicolumn{3}{c}{480p} & \multicolumn{3}{c}{Full Resolution (4K)} \\
        \cline{3-8}
        & & PSNR & SSIM & $\Delta E_{ab}$ & PSNR & SSIM & $\Delta E_{ab}$ \\
        \midrule[1pt]
        % Dis-Rec~\cite{CVPR18DisRec} & 259.2M & 21.98 & 0.856 & 10.42 & 21.81 & 0.862 & 10.60 \\
        UPE~\cite{CVPR19UPE} & 927.1K & 21.88 & 0.853 & 10.80 & 21.65 & 0.859 & 11.09 \\
        DPE~\cite{CVPR18DPE} & 3.4M & 23.75 & 0.908 & 9.34 & - & - & - \\
        HDRNet~\cite{TOG17HDRNet} & 483.1K & 24.66 & 0.915 & 8.06 & 24.52 & 0.921 & 8.20 \\
        % DeepLPF~\cite{CVPR20DeepLPF} & 1.7M & 24.73 & 0.916 & 7.99 & - & - & - \\
        CSRNet~\cite{ECCV20CSRNet} & 36.4K & 25.17 & 0.921 & 7.75 & 24.82 & 0.924 & 7.94 \\
        3D-LUT~\cite{TPAMI203DLUT} & 593.5K & 25.29 & 0.920 & 7.55 & 25.25 & 0.930 & 7.59 \\
        SA-3DLUT~\cite{ICCV21SALUT}* & 4.5M & \textcolor{red}{25.50} & / & / & / & / & / \\
        Ours-S & 47.2K & 25.42 & \textcolor{blue}{0.921} & \textcolor{red}{7.51} & \textcolor{blue}{25.40} & \textcolor{blue}{0.931} & \textcolor{red}{7.52} \\
        Ours-L & 119.8K & \textcolor{blue}{25.47} & \textcolor{red}{0.921} & \textcolor{blue}{7.54} & \textcolor{red}{25.43} & \textcolor{red}{0.932} & \textcolor{blue}{7.56} \\
        \bottomrule[1pt]
    \end{tabular}
    \label{tab:exp-sota-fivek-srgb}
\end{table}

\begin{table}[t]
    \begin{minipage}{0.42\linewidth}
        \centering
        \caption{Quantitative comparisons on the \textbf{FiveK} dataset (480p)~\cite{CVPR11FiveK} for the \textbf{tone mapping}.}
        \setlength{\tabcolsep}{3pt}
        \begin{tabular}{cccc}
            \toprule[1pt]
            \multirow{2}{*}{Method} & \multicolumn{3}{c}{480p} \\
            \cline{2-4}
                                    & PSNR   & SSIM   & $\Delta E_{ab}$  \\
            \midrule[1pt]
            UPE~\cite{CVPR19UPE} & 21.56 & 0.837 & 12.29 \\
            DPE~\cite{CVPR18DPE} & 22.93 & 0.894 & 11.09 \\
            HDRNet~\cite{TOG17HDRNet} & 24.52 & 0.915 & 8.14 \\
            CSRNet~\cite{ECCV20CSRNet} & 25.19 & 0.921 & 7.63 \\
            3D-LUT~\cite{TPAMI203DLUT} & 25.07 & 0.920 & 7.55 \\
            Ours-S & \textcolor{blue}{25.42} & \textcolor{blue}{0.920} & \textcolor{blue}{7.43} \\
            Ours-L & \textcolor{red}{25.43} & \textcolor{red}{0.922} & \textcolor{red}{7.43} \\
            \bottomrule[1pt]
        \end{tabular}
        \label{tab:exp-sota-fivek-xyz}
    \end{minipage}
    \hspace{1pt}
    \begin{minipage}{0.54\linewidth}
        \centering
        \caption{Quantitative comparisons on the \textbf{PPR10K} dataset (360p)~\cite{ICCV21PPR10K} for \textbf{photo retouching}, where a, b, and c denote the groundtruths retouched by three experts} %"$\dagger$" indicates models with ResNet-18 backbone.}
        \setlength{\tabcolsep}{3pt}
        \begin{tabular}{ccccc}
            \toprule[1pt]
            GT & Metric & 3D-LUT\cite{TPAMI203DLUT} & Ours-S & Ours-L \\
            \midrule[1pt]
            \multirow{2}{*}{a} & PSNR & 25.64 & \textcolor{blue}{26.19} & \textcolor{red}{26.28} \\
            & $\Delta E_{ab}$ & 6.96 & \textcolor{blue}{6.71} & \textcolor{red}{6.59} \\
            \midrule[1pt]
            \multirow{2}{*}{b} & PSNR & 24.70 & \textcolor{blue}{25.17} & \textcolor{red}{25.23} \\
            & $\Delta E_{ab}$ & 7.71 & \textcolor{blue}{7.50} & \textcolor{red}{7.49} \\
            \midrule[1pt]
            \multirow{2}{*}{c} & PSNR & 25.18 & \textcolor{blue}{25.51} & \textcolor{red}{25.59} \\
            & $\Delta E_{ab}$ & 7.58 & \textcolor{red}{7.48} & \textcolor{blue}{7.51} \\
            \bottomrule[1pt]
        \end{tabular}
        \label{tab:exp-sota-ppr10k}
    \end{minipage}
\end{table}

\begin{figure}[t]
    \centering
    \includegraphics[width=\linewidth]{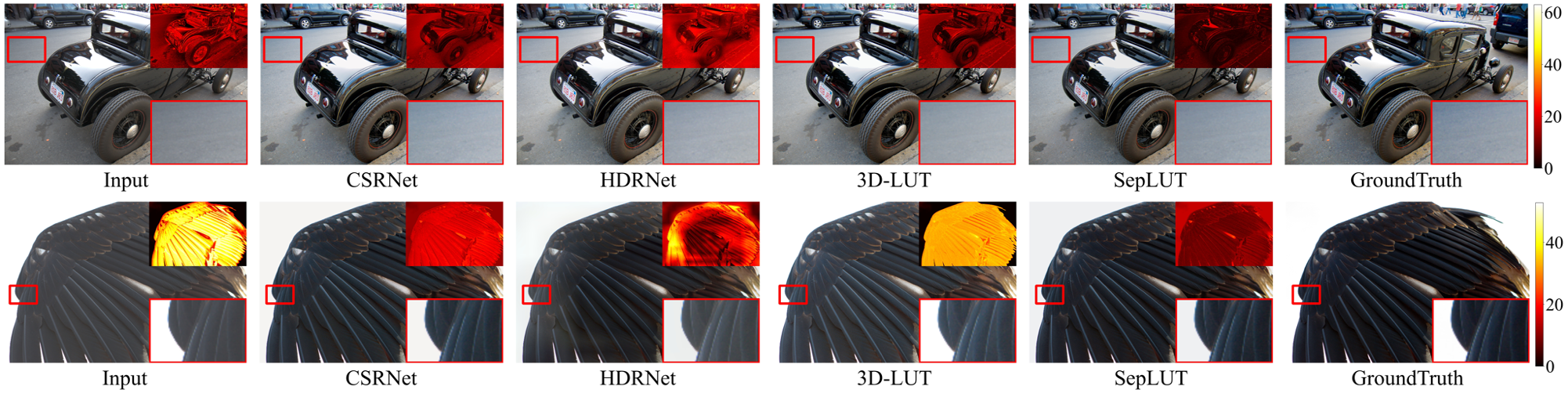}
    \caption{Qualitative comparisons for \textbf{photo retouching} on the \textbf{FiveK} dataset (4K)~\cite{CVPR11FiveK}. The corresponding error maps are placed at the top-right of each image, where brighter colors indicate larger errors. Best viewed on screen.}
    \label{fig:qualitative}
\end{figure}

\subsubsection{Quantitative and Qualitative Comparisons}
% We compare the proposed methods with \textit{state-of-the-art} real-time photo enhancement methods.
For comparisons with \textit{state-of-the-art} real-time methods, we select two typical settings for our approach, namely $m=6, S_o=S_t=9$ and $m=8, S_o=S_t=17$, denoted as \textit{Ours-S} and \textit{Ours-L}, respectively.
\Cref{tab:exp-sota-fivek-srgb,tab:exp-sota-fivek-xyz,tab:exp-sota-ppr10k} report the quantitative comparisons in terms of PSNR, SSIM \cite{SSIM}, and the $L_2$-distance in CIE LAB color space ($\Delta E_{ab}$). The results of the selected methods are obtained via using their publicly available codes and default configurations. The proposed method outperforms others considerably with even fewer parameters. Note that in \Cref{tab:exp-sota-fivek-srgb}, SA-3DLUT \cite{ICCV21SALUT} achieves slightly better performance than our method but at the cost of a significant model size increase (about 37 times) and speed decrease (about 3 times, see \Cref{tab:sota-time}). We also provide some visual comparisons in \Cref{fig:qualitative}, where our method produces more visually pleasing results than others. For example, while the enhanced images of other methods suffer from incorrect brightness or hazy ill-effects, those of our approach present enhanced contrast and sufficient saturation. Please refer to the supplementary materials for more qualitative results.

\subsubsection{Real-time Performance Comparisons}
To demonstrate the practicality of the proposed method, we evaluate the inference time on 100 images and report the average. Each image is tested 100 times on different resolutions including 480P (640$\times$480), 720P (1280$\times$720), 4K (3840$\times$2160) and 8K (7680$\times$4320). The time measure is conducted on a machine with an Intel(R) Xeon(R) Platinum 8163 CPU and an NVIDIA Tesla V100 GPU. As listed in \Cref{tab:sota-time}, our method exceeds the requirement of real-time processing by a large margin on both GPUs and CPUs. The high efficiency of our approach mainly benefits from two factors. First, the downsampled, fixed-resolution input fed to the CNN network makes its computational cost fixed and insensitive to the input resolution. Second, the LUT transform is highly efficient as it is parallelizable on GPUs and can benefit from the fixed-point arithmetic on CPUs.

\begin{table}[t]
    \centering
    \caption{Running time (in millisecond) comparisons on different input resolutions. "-" means the result is not available due to insufficient GPU memory. The "*" symbol indicates that the results are adopted from the original paper (some are absent ("/")) due to the unavailable source code.}
    \setlength{\tabcolsep}{10pt}
    \begin{tabular}{cccccc}
        \toprule[1pt]
        \multirow{2}{*}{Resolution} & \multicolumn{4}{c}{GPU} & CPU \\
         & 480P & 720P & 4K & 8K & 480P \\
        \midrule[1pt]
        % DeepLPF \cite{CVPR20DeepLPF} & 32.12 & 81.23 & - & - & 785.01 \\
        UPE \cite{CVPR19UPE} & 4.27 & 6.77 & 56.88 & 249.77 & 126.37 \\
        DPE \cite{CVPR18DPE} & 7.21 & - & - & - & 553.36 \\
        HDRNet \cite{TOG17HDRNet} & 3.49 & 5.59 & 56.07 & 241.90 & 130.94 \\
        CSRNet \cite{ECCV20CSRNet} & 3.09 & 8.80 & 77.10 & 308.78 & 349.61 \\
        SA-3DLUT \cite{ICCV21SALUT}* & 2.27 & 2.34 & 4.39 & / & / \\
        3D-LUT \cite{TPAMI203DLUT} & 1.02 & 1.06 & 1.14 & 2.35 & 15.52 \\
        Ours-S & 1.08 & 1.09 & 1.18 & 2.23 & 15.65 \\
        Ours-L & 1.10 & 1.12 & 1.20 & 2.35 & 16.21 \\
        \bottomrule[1pt]
    \end{tabular}
    \label{tab:sota-time}
\end{table}
\section{Conclusion}

In this paper, we present a novel framework called SepLUT that simultaneously takes advantage of two different types of LUTs, both 1D and 3D, for real-time image enhancement. It separates a single color transform into component-independent and component-correlated sub-transforms. Extensive experiments demonstrate that such a scheme helps sufficiently exert the capabilities of both types of LUTs and presents several promising properties, including enhanced expressiveness, high efficiency, and light memory footprints. The feasibility of the proposed method reflects that the principle of divide-and-conquer can reduce the capability requirements and ease the optimization of each sub-module, which can significantly increase efficiency. Besides, a proper module decomposition can also benefit from the capability complement between the sub-modules and even achieve enhanced overall performance.

\noindent \textbf{Acknowledgements.}
Yi Xu is supported in part by National Natural Science Foundation of China (62171282, 111 project BP0719010, STCSM 18DZ2270700) and Shanghai Municipal Science and Technology Major Project (2021SHZDZX0102).

\clearpage
\bibliographystyle{splncs04}
\bibliography{egbib}

\appendix

\section{Additional Visual Analysis}

\begin{figure}[t]
    \centering
    \includegraphics[width=0.98\linewidth]{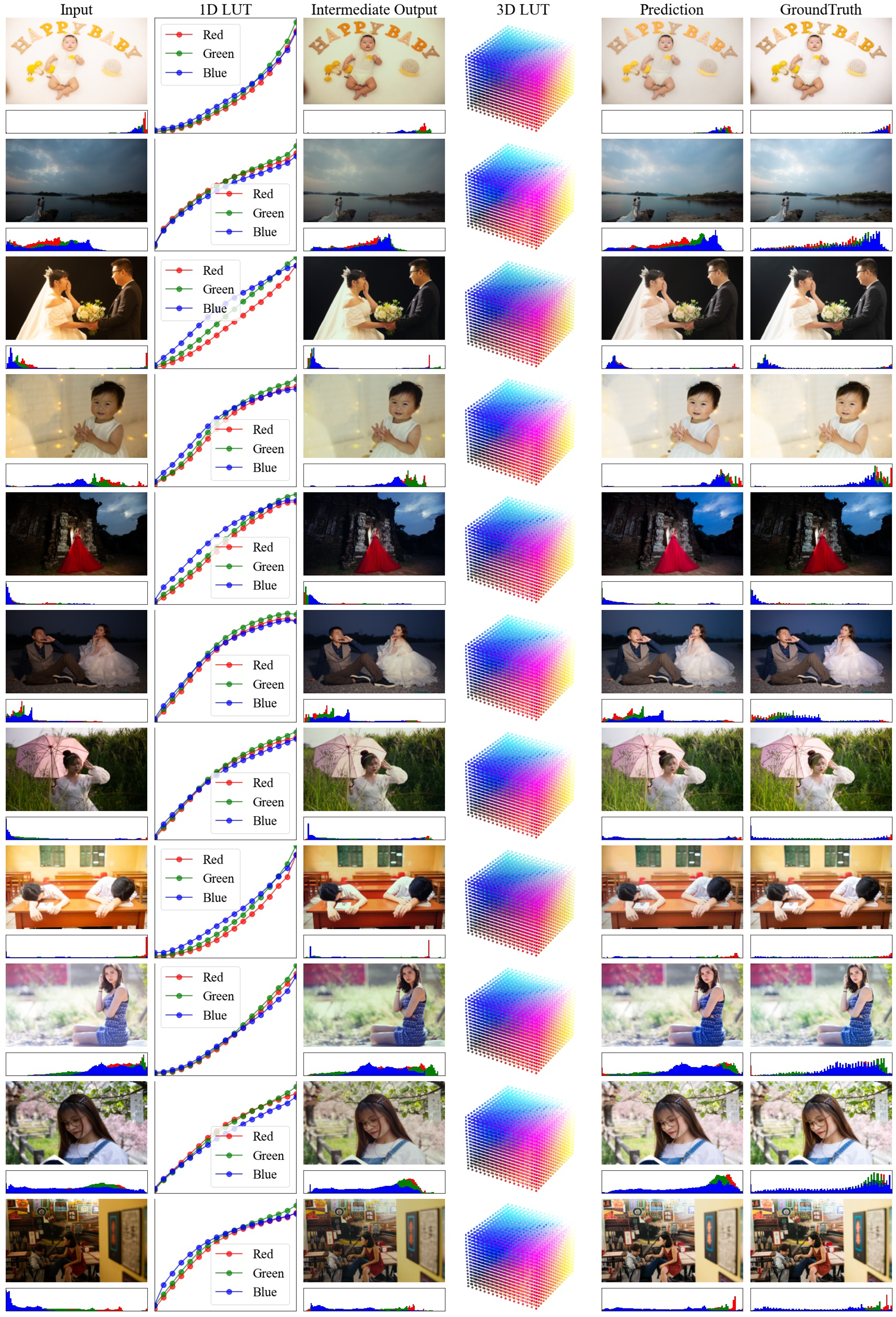}
    \caption{Illustration of the transform procedure and the predicted LUTs. Images are selected from the PPR10K dataset (360P). Best viewed on screen.}
    \label{fig:procedure-ppr10k}
\end{figure}

As shown in \Cref{fig:procedure-ppr10k}, the predicted 1D LUT in the proposed method is mainly responsible for adjusting the brightness and contrast of the input image in an image-adaptive manner, making its histogram more similar to that of the groundtruth. To verify this, we compare the histogram \textit{chi-square distance} \cite{pele2010quadratic} between the input image and the groundtruth with the distance between the image transformed by the 1D LUT and the groundtruth. The chi-square distance is a typical measure of dissimilarity between two histograms, formulated as
\begin{equation}
    \chi^2_{A,B}=\sum_{i=1}^{N}\frac{(x_i-y_i)^2}{x_i+y_i},\ A=\{x_i\}_{i=1}^{N},\ B=\{y_i\}_{i=1}^{N}.
\end{equation}
The statistics show that the predicted 1D LUT decreases the average histogram chi-square distance on the FiveK dataset from 1.6895 to 0.7520, with a 55.49\% reduction. The result is in line with our expectations and is also consistent with the observation in our visual analysis.

\section{Additional Qualitative Results}

% In this section, we provide additional visual comparisons on the FiveK (4K) dataset in \Cref{fig:sota-fivek} and on the PPR10K (360p) dataset in \Cref{fig:sota-ppr10k-1}.

\begin{figure}[t]
    \centering
    \includegraphics[width=0.98\linewidth]{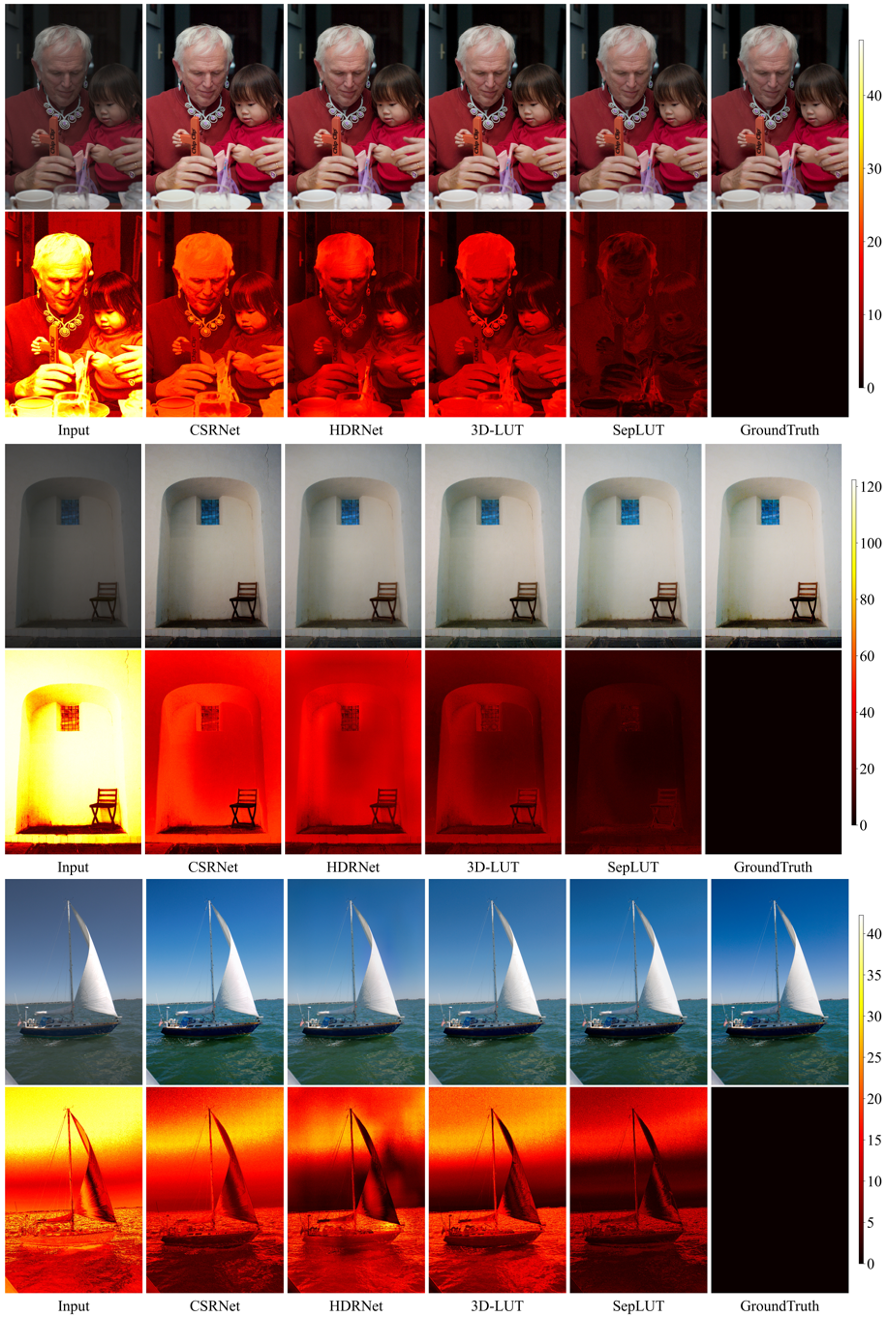}
    \caption{Qualitative comparisons with corresponding error maps on the \textbf{FiveK} dataset (4K) for the \textbf{photo retouching} task. Best viewed on screen.}
    \label{fig:sota-fivek}
\end{figure}

\begin{figure}[t]
    \centering
    \includegraphics[width=0.98\linewidth]{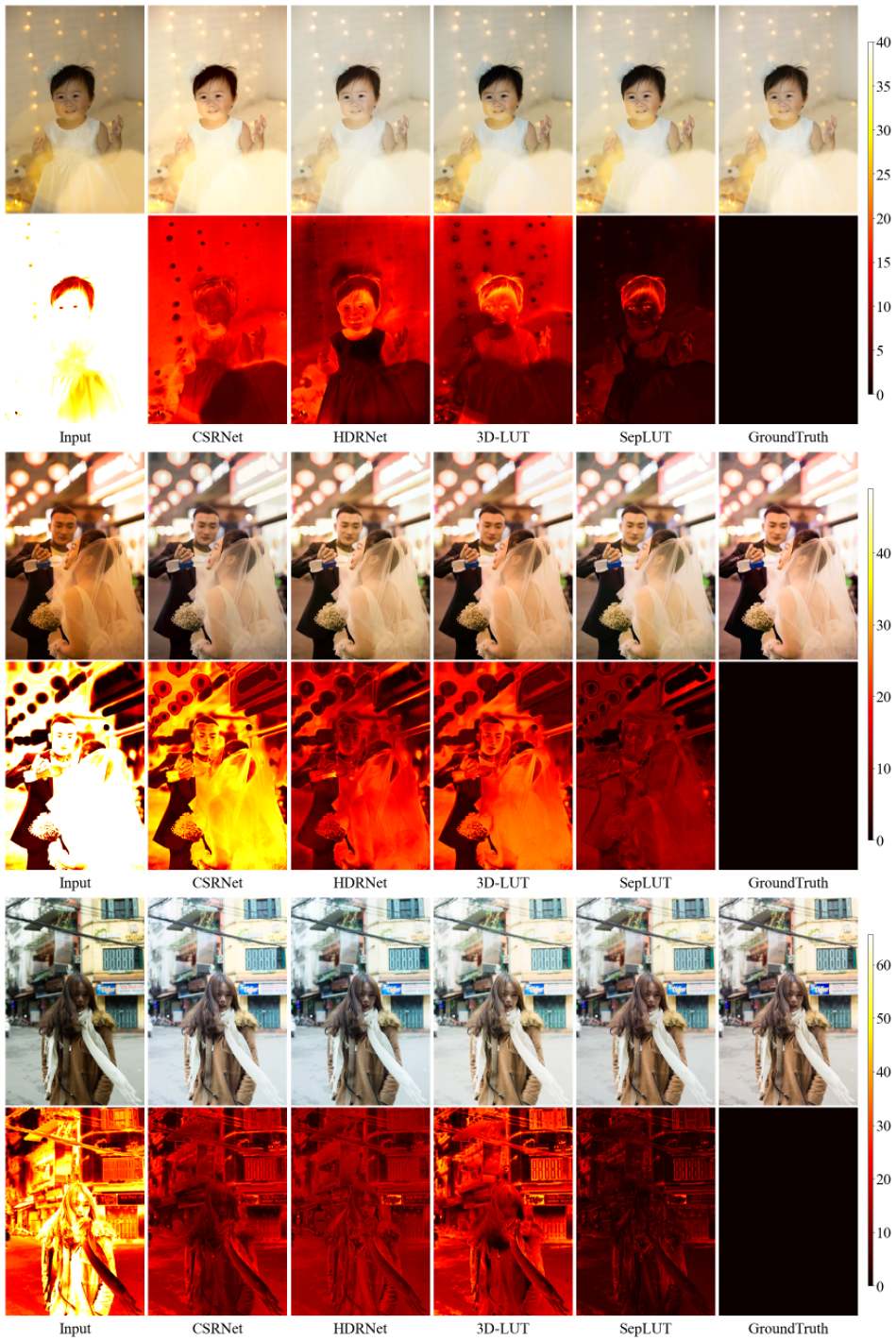}
    \caption{Qualitative comparisons with corresponding error maps on the \textbf{PPR10K} dataset (360P) for the \textbf{photo retouching} task. Best viewed on screen.}
    \label{fig:sota-ppr10k-1}
\end{figure}

In this section, we provide additional visual results in \Cref{fig:sota-fivek,fig:sota-ppr10k-1}.

% \begin{figure}[t]
%     \centering
%     \includegraphics[width=0.98\linewidth]{figures/suppl/Procedure-Visualization-FiveK-1.png}
%     \caption{Illustration of the transform procedure and the predicted LUTs. Images are selected from the FiveK dataset (480P). Best viewed on screen.}
%     \label{fig:procedure-fivek}
% \end{figure}

% \begin{figure}[t]
%     \centering
%     \includegraphics[width=0.98\linewidth]{figures/suppl/Qualitative-Comparison-PPR10K-2.png}
%     \caption{Qualitative comparisons with corresponding error maps on the \textbf{PPR10K} dataset (360P) \cite{ICCV21PPR10K} for the \textbf{photo retouching} task. Best viewed on screen.}
%     \label{fig:sota-ppr10k-2}
% \end{figure}

% \begin{figure}[t]
%     \centering
%     \includegraphics[width=0.98\linewidth]{figures/suppl/Qualitative-Comparison-PPR10K-3.png}
%     \caption{Qualitative comparisons with corresponding error maps on the \textbf{PPR10K} dataset (360P) \cite{ICCV21PPR10K} for the \textbf{photo retouching} task. Best viewed on screen.}
%     \label{fig:sota-ppr10k-3}
% \end{figure}

\end{document}